# Online NEAT for Credit Evaluation - a Dynamic Problem with Sequential Data

Yue Liu
Southern University of Science and Technology
Shenzhen, China
liur3@mail.sustech.edu.cn

Adam Ghandar
Southern University of Science and Technology
Shenzhen, China
aghandar@sustech.edu.cn

Georgios Theodoropoulos
Southern University of Science and Technology
Shenzhen, China
theogeorgios@gmail.com

## ABSTRACT
In this paper, we describe application of Neuroevolution to a P2P lending problem in which a credit evaluation model is updated based on streaming data. We apply the algorithm Neuroevolution of Augmenting Topologies (NEAT) which has not been widely applied generally in the credit evaluation domain. In addition to comparing the methodology with other widely applied machine learning techniques, we develop and evaluate several enhancements to the algorithm which make it suitable for the particular aspects of online learning that are relevant in the problem. These include handling unbalanced streaming data, high computation costs, and maintaining model similarity over time, that is training the stochastic learning algorithm with new data but minimizing model change except where there is a clear benefit for model performance.

## 1. INTRODUCTION

Credit score evaluation to determine trust and reliability of partners in financial transactions is a key issue for the microfinance industry. The development of peer to peer (P2P) lending provides a mechanism for wider inclusion and participation in the financial economy [1]. The complex background of unbanked applicants poses challenges for these individuals' access to finance and banking (including saving, borrowing to support a business or cushion against emergencies). The Chinese online P2P lending industry which first emerged in 2007 is a case in point. The total number of online P2P lending platforms reached its peak in 2015, and the figure saw a steady decline despite increasing in demand as small P2P lending companies encountered problems in finance and a significant issue was the rate of defaults arising from an inability to predict the credit worthiness of applicants [2].

These challenges created a need to develop new ways to evaluate trust beyond the traditional approaches utilized in established lending markets. Innovations discussed in the literature that provide ways to address these challenges include utilizing additional data beyond the information used in traditional credit scoring in established markets and applying data mining and learning to develop adaptive models [3, 4]. We find that a crucial aspect is consideration of dynamic patterns in models relating trust and changing information. The result is a significantly more challenging variation of the traditional credit evaluation problem that has been studied previously in the classification and machine learning literature [5, 6].

In this paper, we describe development and application of a technique for learning online (or frequently updated) credit scoring models as new data is read record by record. We describe the approach developed as Online NEAT for Credit Scoring. The approach applies neuro-evolution, a technique that combines neural networks with evolutionary computation [7]. Neuroevolution is a way to optimize neural network models and unlike the widely used gradient descent methods determine the weights and the structure of the neural network model simultaneously [7]. Machine learning methods including neuroevolution are stochastic learning algorithms and one drawback of such methods is that models can change simply because a new local optimum is identified (which may not have advantages in prediction and there may be a business cost in changing the credit evaluation model frequently). In our approach, we use the basis of the technique in a population-based evolutionary algorithm to maintain a population seeded with historically high performing models. Another issue we address is class imbalance in the streaming data. Common approaches to handle unbalanced data in classification include resampling, adjusting the loss function, and using ensembles. Resampling is challenging in online learning where new samples arrive with unbalanced distributions. We describe an adjustment to NEAT where the fitness function is adjusted to give higher penalty to misclassifying the minority (default) class, based on the method presented in [8].

Neuroevolution has demonstrated high performance in a wide range of classification and learning tasks, but it has not been widely applied in credit scoring in the literature. In this paper, we provide a feasible and systematic method to build a credit scoring prediction model with neuroevolution and also compare the performance of proposed strategy with other machine learning methods in on online learning task based on real sequential loan data from peer to peer lending[1] of loan performance.

This remainder of this paper is organized as follows: a literature review is given on neuroevolution and its application to credit scoring in section 2; section 3 describes the proposed approach; section 4 provides empirical analysis; and finally section 4 concludes the paper.

---
[1] https://www.lendingclub.com/info/download-data.action

## 2. LITERATURE REVIEW

The purpose of credit risk evaluation is to develop a classification model that accurately distinguish good applicants from bad applicants that tend to default [9]. Credit decisions for loans can be analyzed based on 5 factors [10] to assess loan quality:

- Character or reputation of the borrower or applicant.
- Capital or leverage defined as the amount of debt the borrower is already responsible for.
- Capacity to repay defined as volatility of the borrowers' income.
- Collateral used as security for the loan.
- Cycle which is the current stage of the economic cycle (a function of macroeconomic conditions).

These factors are termed the 5 C's, the cycle is clearly a dynamic quantity and the other factors also may exhibit dynamic relationships with ability to repay debt. Other research has also demonstrated the use of dynamic models and the variable behavior of borrowers and static approaches have been shown to be outperformed by dynamic scoring approaches in other research [11].

Many statistical and machine learning approaches have been applied into credit scoring to improve prediction accuracy. An overview includes: Logistic Regression [12], K-nearest Neighbours [13], Decision Trees [14], Support Vector Machines [15, 16], Neural Network [10,17]. Hybrid techniques that combine different classification techniques together have demonstrated strong performance [18, 19]. Ensemble techniques that use variations of model structure and data and combine diverse models to obtain a prediction have been shown to be particularly effective [20].

The majority of studies in the literature concentrate on reporting improvements in prediction performance (classification accuracy or sometimes precision and recall due to the issue of unbalanced classes). Other attributes of performance that may be of interest include computation cost, interpretability and use of real data / different problem variations are often not a key focus. In this paper, we compare and analyze these aspects of performance also. Another area of our contribution is in the consideration of the dynamic aspects of the problem and online learning. In real problems, especially in P2P lending applications, large volumes of data for applicants including new user data and information are added daily or more frequently. The dynamic aspects of the patterns required for lending decisions discussed previously mean that static models could be inefficient or even detrimental to performance.

Artificial Neural Networks (NN) have been shown to be effective for many types of supervised learning problems. However, parameters including the structure of the network, the learning samples and the initial values have a large impact on the final performance [21]. Evolutionary Algorithms (EAs), inspired by natural selection and genetics, are a strategy that has been used to optimize the design of neural networks. An obvious advantage of Evolutionary Algorithms (EA) is that EA is a global search strategy that can provide benefits in avoiding local optima during the searching process without placing strong requirements on the problem representation [22] such as convexity.

These advantages have led to a wide spectrum of evolutionary approaches for optimizing the design of neural networks. In the beginning, many studies focused on evolving small and fixed-topology networks [23]. In 2002, NeuroEvolution of Augmenting Topologies (NEAT) was proposed and became one of the most widely applied neuroevolution models [24]. The main drawback of the approach is scalability to larger models. There are a number of algorithms extending NEAT which use different schemes to learn large networks, for instance, HyperNEAT attempted to adapt the representation to the task in a way analogous to the use of domain knowledge [25]. DeepNEAT [26] also extends NEAT along these lines while maintaining the inherent concepts of NEAT by a hierarchical/meta optimization process that includes operations on layers (implied by parameters at the nodes) rather than individual neurons. CoDeepNEAT [26], is a related approach using cooperative co-evolution to learn deep networks in where modules which were put together by a design plan.

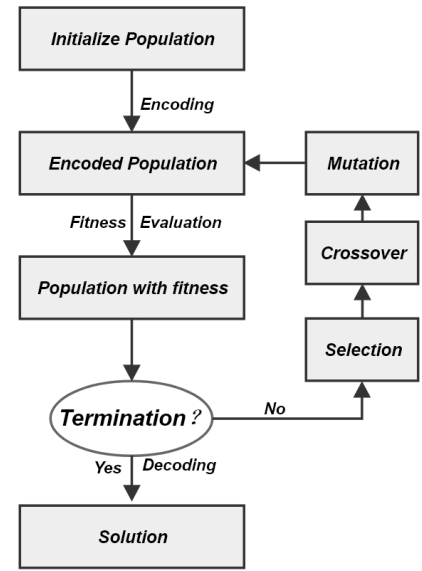
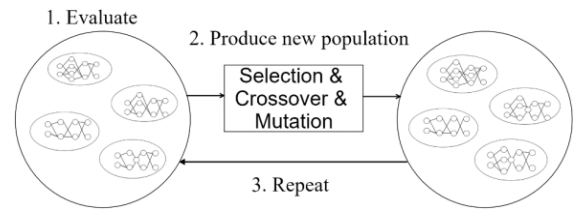

**Figure 1. Evolution algorithm**

The benefits of Neuroevolution as an approach for optimizing NN compared with other common ML methods is demonstrated by Miguel and Paulo [27] where they showed NEAT could obtain competitive results on sixteen real-world tasks of classification and regression compared with a set of other data mining algorithms [27]. For another recent application where NEAT outperformed other methods in a real-world classification problem, see [28]. This paper applies neuroevolution to credit evaluation formulated as a classification problem and discusses the potential value. More than that, we will discuss the capability of NEAT when dealing with sequential data, which is more familiar to the application scenarios in the real world.

## 3. METHODOLOGY

In this section, we discuss the implementation framework including Evolutionary algorithm, NEAT, and NN representation as well as a formulation of the credit scoring problem for loan profitability.

## 3.1 Evolutionary Algorithm

Evolutionary algorithms are a computational problem-solving technique inspired by natural evolution and genetics. The family of algorithms are global meta-heuristics that can avoid many of the problems of local search techniques on different search spaces [29].

A Genetic algorithm has four main components: a way of *representing* and encoding the problem; a method of initializing a *population* of individuals represented as such; a *fitness* function to *evaluate* each individual in the population; and *operators* for variation of the individuals to simulate producing offspring, crossover and mutation [30].

The general process of a genetic algorithm is shown in Fig.1. By fitness evaluation, the algorithm will evaluate all the solutions in the population and eliminate some poor performing individuals from the population. The power of the intrinsic parallelism of genetic search is amplified by the mechanics of population modification, allowing the genetic algorithms to attack even NP-hard problems [31].

## 3.2 Neuroevolution of Augmenting Topologies (NEAT)

The core concept of Neuroevolution is using evolutionary computation to optimize the structure and weights of neural network. A basic design of neuroevolution is shown in Figure 2. NEAT is one of the most popular neuroevolution methods, which realizes the evolution simultaneously of the topology and weights of neural networks. This subsection briefly reviews the NEAT method, see also [24] for a comprehensive introduction.

NEAT uses a flexible approach for genetic encoding, which allows topology represented by the genome to smoothly change with the application of genetic operators. Each genome includes a list of connection genes that store the related information about genes connection, including inputs, outputs, weights, an enable bit and an innovation number. Importantly NEAT maintains a record of the generation in which parts of the genomes were formed to maintain a meaningful relationship between different parts of the structure (i.e. different neurons and connections work together).

In order to determine exactly which gene matches up with which, NEAT introduce historical markings to track the historical origin of genes. When a new gene is produced during the evolutionary process, a global innovation number will be created and assigned to the gene to record the historical information. Based on the innovation numbers, genes are able to judge the origin and topological similarity and know exactly which genes match up with which. In order to preserve diversity, speciation is introduced into NEAT, so as to protect innovation. Another innovation that reduces the computation complexity of NEAT compared with other neuroevolutionary techniques is by initializing the population with simpler NN genotypes (no hidden layers) and adding complexity where a benefit is identified by fitness evaluation. The advantage

```
1: function ONLINENEAT(TimeWindowSize, Data)
2:   N ← len(Data)/TimeWindowSize;
3:   Split the Data into N parts based on time sequence, {data_1,data_2,...,data_N};
4:   Generate initial individuals in the population;
5:   for i = 1 → N do
6:     CurrentTimeWindow ← data_i;
7:     while Termination not satisfied do
8:       Select individuals from the population;
9:       Crossover and mutation operations;
10:      Add new individuals into offsprings;
11:      Evaluate the fitness of individuals(neural networks) in the offsprings
12:          and population based on CurrentTimeWindow;
13:      Replace least-fit individuals in the population
14:          with offsprings with high fitness;
15:    end while
16:  end for
17: end function
```

**Algorithm 1. Online Neat algorithm**

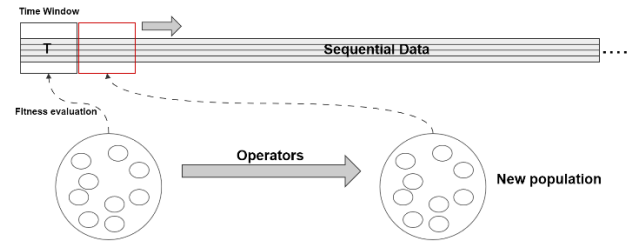

**Figure 2. Online Neat algorithm**

of this design is that the algorithm doesn't need to optimize complex network structures that have limited value and instead can focus on exploration of useful structures. In this way, NEAT enables evolution to be applied to both optimize and complexify solutions simultaneously.

## 3.3 Online NEAT for Credit Risk Evaluation

There is no systemic and integrated research focused on utilizing the particular characteristics of neuroevolution to improve the performance of credit scoring models. We propose here an online version of NEAT which includes credit risk analysis and time series data.

### 3.3.1 Online NEAT for Dynamic Classification of Credit Default Risk

In real-world P2P credit score evaluation, the volume of data is high and the environment in which decisions take place is dynamic. Models and data can be updated at frequent intervals (daily, hourly or shorter depending on the number of customers). However, it is observed that in the literature applying machine learning to credit scoring that by far the most common method to evaluate credit risk is using randomly selected samples to train a model and then apply it in testing in out of sample data points[2]. This is analogous to a business model for credit scoring where a static model is created to deal with new applications and never updated. Or rebuilding a new

---

[2] A selection of datasets for credit evaluation used in the literature widely used in ML research. None of the datasets includes information on the date applications were made so is not possible to evaluate time sensitivity of patterns found.

- https://archive.ics.uci.edu/ml/datasets/statlog+(german+credit+data)
- https://archive.ics.uci.edu/ml/datasets/credit+approval
- https://archive.ics.uci.edu/ml/datasets/default+of+credit+card+clients
- http://archive.ics.uci.edu/ml/datasets/statlog+(australian+credit+approval)

model after a large volume of new data is obtained not incrementally. Consequently, it is not possible to comprehensively evaluate the relation of performance to time and adaptability, scalability and need for efficiency for many proposed models in more realistic scenarios. Using evolutionary algorithms to train models, especially with a huge dataset, can be time-consuming, due to inevitable costs to continuous model building and fitness evaluation on each individual in population during the genetic process. An additional benefit of online neuroevolution for credit risk evaluation is that the learning process is not restarted as new data is read.

The processes of online NEAT are depicted in Figure 2 and Algorithm 1. The proposed algorithm introduces a new concept into the original method: the concept of "Time Window" records the data currently in use. The length is delineated by a number of records. As shown in Figure 2, an individual in the population represents a specific structure of neural network, and the individual will mutate and mate with other individuals, producing new individuals in this population. Because of the fixed size of this population, all individuals compare their fitness values based on the data in the current time window with these new offspring. By comparison, the population retains the best N neural networks that have better predictive performances compared to the eliminated individuals. The processes will continue until the population has found the optimal solution in this time window.

For a current population, all individuals continue to evolve to the best status based on the data in the current time window. If the population has found the optimal solution in this time window, the window will move to the next and the population begins a new round of evolution. In this way, the model became robust during the evolutionary process.

### 3.3.2 Fitness Evaluation and Class Imbalance

Another advantage of evolutionary algorithms is these algorithms can control the direction of evolution (i.e. a search process and preference for solutions) by altering the fitness function. In general, other machine learning methods cannot be so flexible in changing the preference for solutions or incorporating meta concepts about solution characteristics (such as class imbalance, model interpretability, etc), since they have rigid inherent structures that direct the process. In particular, we are concerned with class imbalance as an inevitable problem in realistic application situations, especially in credit risk evaluation, due to a low proportion of default examples compared with non-default examples. A widely used method to cope with imbalance is resampling, converting an imbalance classification problem to a balanced classification, nevertheless, an obvious issue is over-sampling or under-sampling, giving rise to bad performance in final results [32]. However, this is not possible in the case of online learning where the new samples may be arriving with unbalanced distributions. In this case, it has been shown that it is possible to adjust the fitness function to give a preference to identifying the less frequently occurring class for instance (see [8]). We apply this method in our implementation of NEAT for online learning.

**Table 1. Confusion matrix**

|  | Predicted positive | Predicted negative |
|---|---|---|
| **Actual positive** | TP | FN |
| **Actual negative** | FP | TN |

$$Accuracy = \frac{TP+TN}{TP+TN+FT+FN} \quad (1)$$

$$Recall\ (Sensitivity) = \frac{TP}{TP+FN} \quad (2)$$

$$Specificity = \frac{TN}{TN+FT} \quad (3)$$

Accuracy, recall and specificity are metrics used to evaluate 2-class classification performance. Accuracy is a quite common metric of prediction performance in research or in practical applications, which provides the overall accuracy for the whole dataset. However, when there is an imbalance in the data, accuracy cannot evaluate performance very well since a high accuracy can just result from its predictive performance on the major class. Recall (Sensitivity) and Specificity measure the accuracy on positive applications and negative applications respectively, which enable to evaluate a model more comprehensively. Table 1 and Function (1) - (3) show the counting process of these metrics.

The first method to evaluate the fitness is using accuracy (ACC), which can be described as

$$fitness = Accuracy = \frac{TP+TN}{TP+TN+FT+FN} \quad (4)$$

The second one combines Recall and Specificity since the imbalanced proportion of positive and negative (PAN), which can be described as

$$fitness = Recall + Specificity = \frac{TP}{TP+FN} + \frac{TN}{TN+FT} \quad (5)$$

The third one utilizes profits of each loan (PRO), and the empirical formulas are described in Table 2 and Function 6.

**Table 2. Profit matrix** (L: the amount of loan for a borrower, I: total interest from this loan)

| profit(i) | Predicted positive | Predicted negative |
|---|---|---|
| **Actual positive** | I | -I |
| **Actual negative** | -L | L |

$$fitness = profit = \sum_m profit\ (i) \quad (6)$$

Due to making a model based on the sequential data, there may be a close connection of fitness value between individuals in contiguous generations. A new population calculate its fitness based on a new data segment in the next time window, so historical fitness will be introduced (PAP) in last method, which can be described as R

$$fitness(t) = \alpha \cdot profit(t) + \beta \cdot fitness(t-1) \quad (7)$$

## 4. Empirical Results

We use a real-world P2P loan dataset to evaluate the proposed approaches. This section introduces the loan dataset and compares and analyzes the performance of proposed approaches. In our experiments, we consider several variations of dynamic sampling based on time windows and use different fitness evaluation strategies to solve the current issues.

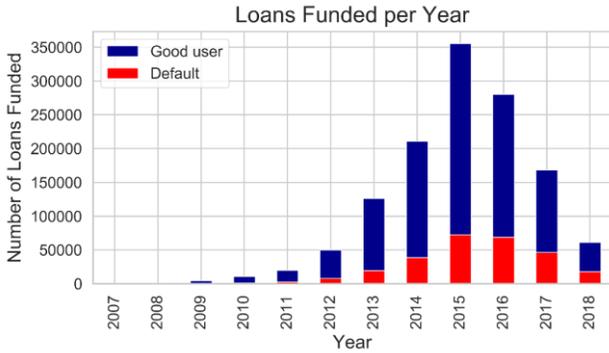

Figure 3. Usable Loan funded per year for Lending Club

### 4.1 Lending Club Data Set

The Lending Club[3] was created in 2007, since then the Lending Club has evolved into the largest P2P loan platform in the United States. The amount of loan of lending club maintain the rising trend all the time and reach 47 billion dollars in the first season of 2019. In order to attract more investors, the company has opened up their data on every loan they have ever issued on their website, which is convenient for researchers and investors to analyze. The paper utilized the data from lending club for evaluation. The company provides a variety of configurations for loans (e.g. single borrower, multiple lenders) and interest rates are dependent on borrower attributes. We use information on interest rates and estimated default probability in the model.

As for the data they have provided on their platform, the statistics were recorded since 2007. The number of borrower variables is 151. Since the company provided all the data, which contains a lot of loans which are up to date on all outstanding payments and the loan are in the lending term, meaning that we could not obtain their final status of loans. Meanwhile, the missing information on some loans is very large, so our experiments deleted them. Figure 3 shows useable data of leading club from 2007 to 2018, and the reason why the figure saw a sharp decline after 2015 is that the proportion of applicants who are using their loans normally since the lending term is not over is very high.

### 4.2 Comparison of Machine Learning Techniques and Online NEAT

In this section, we discuss the effectiveness of online NEAT from several aspects and compare its performance with traditional classification algorithms.

#### 4.2.1 Experiments on traditional Approaches

In this section, four traditional classification approaches would be used to evaluate their ability of prediction performance, including K-Nearest Neighbor (KNN), Decision Tree (DT), Random Forest (RT) and Multi-layer Perceptron (MLP). In the training, we will use the usable data in 2017 as the train set to predict the results in 2018. The specific distribution of training and testing data was presented in Table 3.

Table 3. Details of experimental data

|  | Train (2017) | Test (2018) |
| --- | --- | --- |
| **Overall amount** | 168336 | 61164 |

[3] https://www.lendingclub.com/

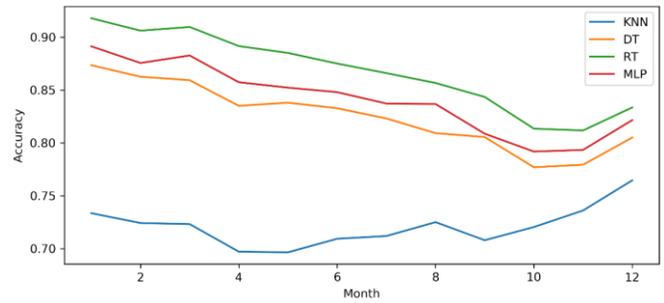

Figure 4. Predictive performance on different months in 2018

Table 4. Traditional ML Models

|  | Accuracy | Recall | Specificity |
| --- | --- | --- | --- |
| **KNN** | 0.716827 | 0.937343 | 0.183043 |
| **DT** | 0.838426 | 0.928898 | 0.619428 |
| **RT** | 0.883502 | 0.984278 | 0.639562 |
| **MLP** | 0.783247 | 0.952002 | 0.374755 |

| **The amount of positive** | 121775 | 43283 |
| --- | --- | --- |
| **The amount of negative** | 46561 | 17881 |
| **Default rate** | 0.2765 | 0.2923 |

Table 4 and Figure 4 show the results of traditional classification methods. From Table 4, we found that all models are able to predict positive applicant well, owing to the high value of recall, but with low specificity. On the other hand, Figure 4 shows the predictive ability on different months in 2018 of the model trained by the data in 2017. An obvious result is that the overall predictive performance saw a stable decline. In detail, the model trained by the data in 2017 could predict the data in January 2018, but the model can not achieve better performance in a longer period of time in the future.

Table 5. Description of methods in the experiments

|  | Description |
| --- | --- |
| **LSTM** | Cost was set as accuracy |
| **ACC+NEAT** | Fitness function (4) |
| **PAN+NEAT** | Fitness function (5) |
| **PRO+NEAT** | Fitness function (6) |
| **PAP1+NEAT** | Fitness function (7), β =0.1 |
| **PAP2+NEAT** | Fitness function (7), β =0.5 |
| **PAP3+NEAT** | Fitness function (7), β =1 |

### 4.2.2 Experiments on online NEAT

The experiments in the last part told us traditional classification algorithms couldn't deal with unbalanced data well, and keeping updating the model is a crucial issue in practical problems. In fact, Long Short-Term Memory (LSTM) is well-suited to classifying, processing and making predictions based on time series data as a recurrent neural network. In this section, we explored the performance of online NEAT on the lending club data and compare its results with LSTM to show its effectiveness on coping with sequential data.

In the experiments, the amount of data we used is 229501 (from 2017 to 2018) and the population size was set to 200. Figure 5 and Figure 6 show the prediction performance for the data in next time window after the end of training on the data in the last time window. Table 5 shows the detailed descriptions of methods in our testing. As for PAP, we set α as 0.000001 to reduce the scale and test three settings on $\beta$, PAP1 ($\beta = 0.1$), PAP2 ($\beta = 0.5$), PAP3 ($\beta = 1$) to assess the impacts of historical fitness. The length of time window was set to 500 and 1000 respectively in Figure 5 and Figure 6. At the same time, we also show the predictive performance of the

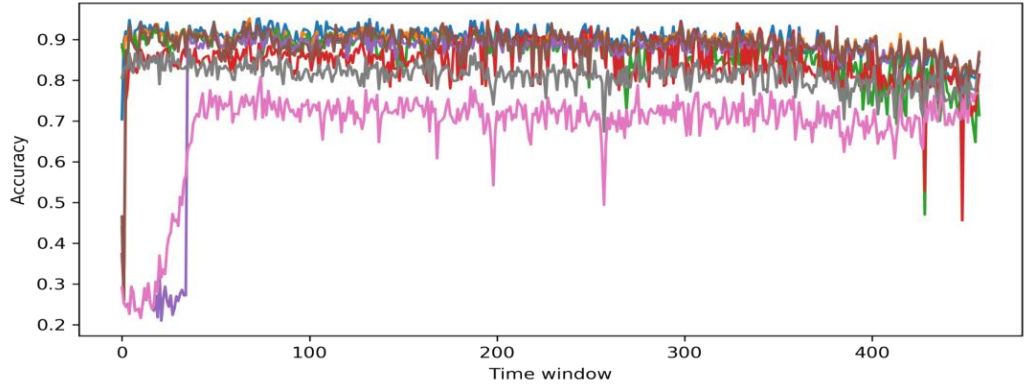

(a). Overall accuracy

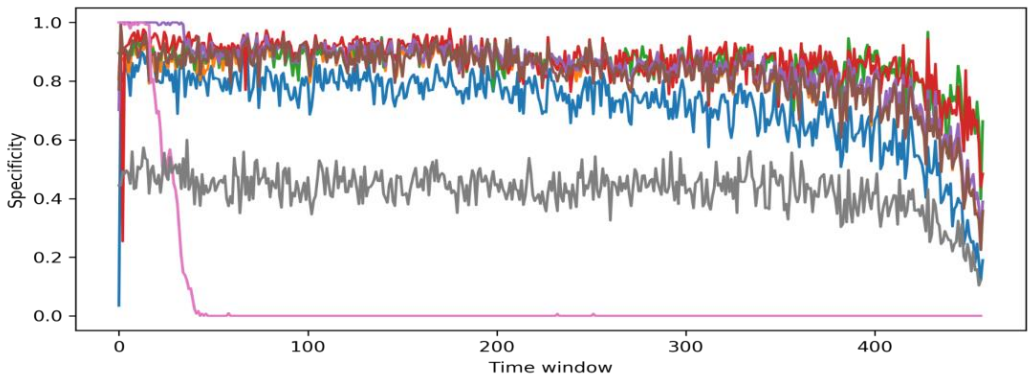

(b) Specificity

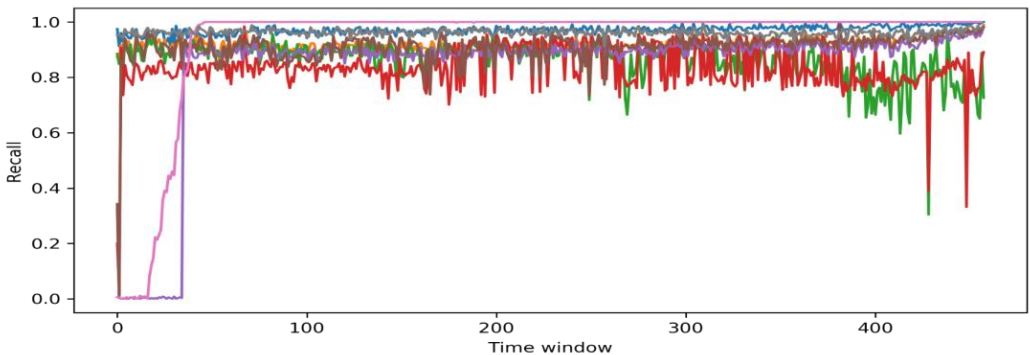

(c) Recall

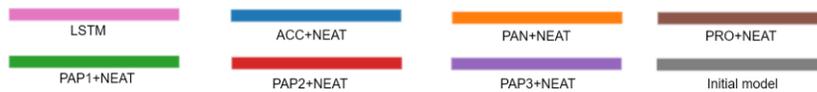

**Figure 5 Prediction performance of Online Neat** (The length of time window is 500)

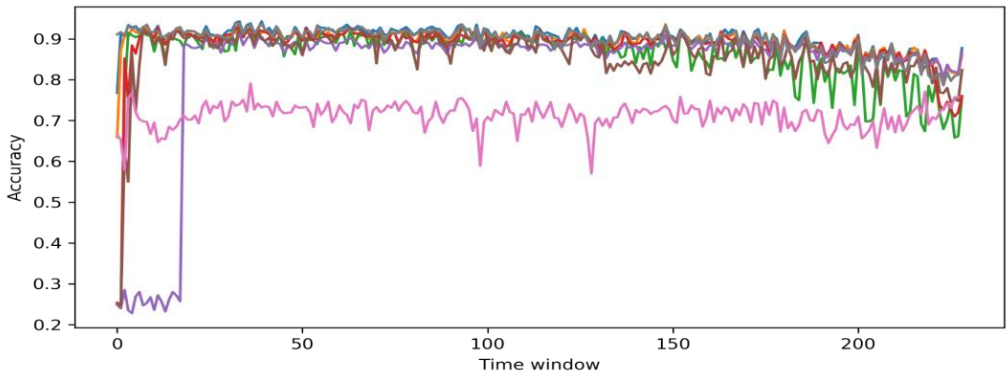

**(a). Overall accuracy**

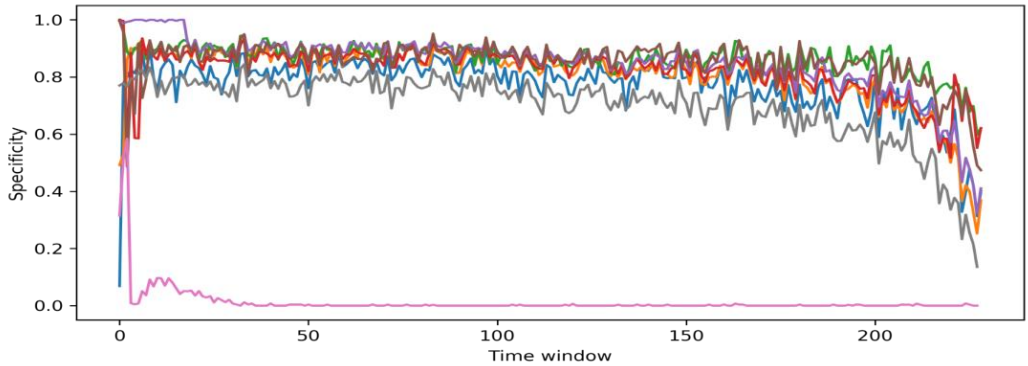

**(b) Specificity**

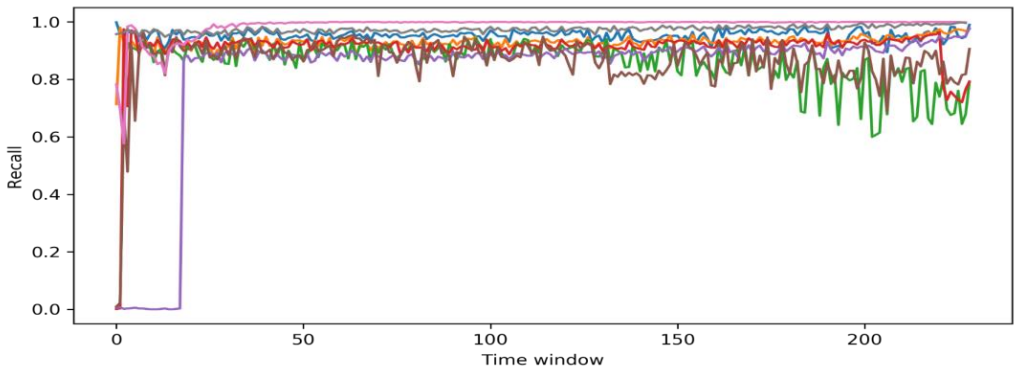

**(c) Recall**

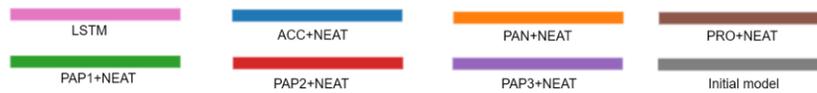

**Figure 6 Prediction performance of Online Neat** (The length of time window is 1000)

initial model built by ACC+NEAT, which was presented in a grey line in Figure 5 and 6.

From the three pictures in Figure 5, we could find that Online Neat has a better performance of classification prediction, because the accuracy of online neat is much higher than LSTM. At the same time, by setting the fitness function specifically, online NEAT had a capability to deal with unbalance and had better adaptability. In detail, the default rate is close to 25% as a whole, and it would change during the process. PAP1+NEAT and PAP2+NEAT can reach better specificity to predict default applications, even though the situation is quite different under a higher proportion of loans in progress in 2018. The performance of the initial neural network told us that NEAT optimized neural networks during the evolution.

From Figure 6, we found a similar result with Figure 5, though the length of time window doubled. LSTM had a bad performance on the accuracy prediction and Online NEAT could achieve better both on specificity and recall. However, there were some differences in the performances of various online NEAT. Although ACC+NEAT had worse specificity compared with other online NEATs, the gap narrowed when the time window largened. On the other hand, when the length of time window was set to 1000, PRO+NEAT and PAP1+NEAT (β=0.1) had a similar performance on all three metrics, but PAP2 (β=0.5) and PAP3 (β=1) achieved better on accuracy and recall. When comparing the performance of the initial model, a bigger time window could improve its performance, but the gap between the initial model and the model after evolution should not be ignored. Overall, online NEAT is able to deal with sequential credit data better than traditional classification methods and LSTM since the adaptability of online NEAT is better.

## 5. CONCLUSION

This paper has presented an online approach for applying neuroevolution for credit risk evaluation. The proposed approach applies NEAT to evolve the predictive model dynamically, so as to improve its performance continually to cope with the sequential data. The online NEAT implementation handles class imbalance by applying an adjustment to the fitness function to consider performance on classifying different classes.

In the experiments we also compare a static model with the dynamic online model and find the dynamic model we develop gives superior performance and it is evident that the environment is changing and the technique provides benefit through adaptation. The online approach was able to predict both positive applications and negative application well even though the default rate is low (classes unbalanced).

In the future, we will continue to improve our proposed approach and deploy it to a real P2P loan server. Another issue is that NEAT is time-consuming when the number of features is large so we will develop methods to handle computation time.